\documentclass[10pt,conference,a4paper]{IEEEtran}
\IEEEoverridecommandlockouts
\usepackage{cite}
\usepackage{amsmath,amssymb,amsfonts}
\usepackage{algorithm}
\usepackage{algorithmic}
\usepackage{graphicx}
\usepackage{textcomp}
\usepackage{xcolor}
\usepackage{caption, threeparttable}
\usepackage{multirow} 
\usepackage{multicol}
\usepackage{booktabs}
\usepackage{subfigure}
\def\BibTeX{{\rm B\kern-.05em{\sc i\kern-.025em b}\kern-.08em
    T\kern-.1667em\lower.7ex\hbox{E}\kern-.125emX}}
\begin{document}

\title{Complementing Representation Deficiency in Few-shot Image Classification: A \\Meta-Learning Approach
\thanks{$^\ast$ Corresponding author}
}
\author{\IEEEauthorblockN{\hspace{-8ex}Xian Zhong}
\IEEEauthorblockA{\textit{\hspace{-8ex}Hubei Key Laboratory of} \\
\textit{\hspace{-8ex}Transportation Internet of Things \&}\\
\textit{\hspace{-8ex}School of Computer Science and}\\
\textit{\hspace{-8ex}Technology,}
\textit{Wuhan University of}\\
\textit{\hspace{-8ex}Technology,}
Wuhan, China \\
\hspace{-8ex}zhongx@whut.edu.cn}
\and
\IEEEauthorblockN{\hspace{-6ex}Cheng Gu}
\IEEEauthorblockA{\textit{\hspace{-6ex}School of Computer Science and} \\
\textit{\hspace{-6ex}Technology,}
\textit{Wuhan University of}\\
\textit{\hspace{-6ex}Technology,}
Wuhan, China \\
\hspace{-6ex}gucheng$\_$hs@whut.edu.cn}
\and
\IEEEauthorblockN{\hspace{-2ex}Wenxin Huang$^\ast$\hspace{4ex}}
\IEEEauthorblockA{\textit{\hspace{-2ex}School of Computer Science\hspace{4ex}} \\
\textit{\hspace{-2ex}Wuhan University\hspace{4ex}}\\
\hspace{-2ex}Wuhan, China\hspace{4ex} \\
\hspace{-2ex}wenxin.huang@whu.edu.cn\hspace{4ex}}
\and
\IEEEauthorblockN{\hspace{3ex}Lin Li}
\IEEEauthorblockA{\textit{\hspace{6ex}School of Computer Science and} \\
\textit{\hspace{6ex}Technology,}
\textit{Wuhan University of}\\
\textit{\hspace{6ex}Technology,}
Wuhan, China \\
\hspace{6ex}cathylilin@whut.edu.cn}
\and
\IEEEauthorblockN{\hspace{6ex}Shuqin Chen}
\IEEEauthorblockA{\textit{\hspace{6ex}School of Computer Science and} \\
\textit{\hspace{6ex}Technology,}
\textit{Wuhan University of}\\
\textit{\hspace{6ex}Technology,}
Wuhan, China \\
\hspace{6ex}csqcwx0801@whut.edu.cn}
\and
\IEEEauthorblockN{\hspace{6ex}Chia-Wen Lin}
\IEEEauthorblockA{\textit{\hspace{6ex}Department of Electrical Engineering} \\
\textit{\hspace{6ex}\& Institute of Communications}\\
\textit{\hspace{6ex}Engineering, National Tsing Hua}\\
\textit{\hspace{6ex}University,}
Hsinchu, Taiwan \\
\hspace{6ex}cwlin@ee.nthu.edu.tw}
}

\maketitle

\begin{abstract}
Few-shot learning is a challenging problem that has attracted more and more attention recently since abundant training samples are difficult to obtain in practical applications. Meta-learning has been proposed to address this issue, which focuses on quickly adapting a predictor as a base-learner to new tasks, given limited labeled samples. However, a critical challenge for meta-learning is the representation deficiency since it is hard to discover common information from a small number of training samples or even one, as is the representation of key features from such little information. As a result, a meta-learner cannot be trained well in a high-dimensional parameter space to generalize to new tasks. Existing methods mostly resort to extracting less expressive features so as to avoid the representation deficiency. Aiming at learning better representations, we propose a meta-learning approach with complemented representations network (MCRNet) for few-shot image classification. In particular, we embed a latent space, where latent codes are reconstructed with extra representation information to complement the representation deficiency. Furthermore, the latent space is established with variational inference, collaborating well with different base-learners, and can be extended to other models. Finally, our end-to-end framework achieves the state-of-the-art performance in image classification on three standard few-shot learning datasets.
\end{abstract}


%
\IEEEpeerreviewmaketitle

\section{Introduction}
Humans have extraordinary abilities to utilize the previous knowledge to quickly learn new concepts from limited information. In contrast, although deep learning methods have achieved great success in many fields, such as image classification, natural language processing, and speech modeling~\cite{SimonyanZ14a, DBLP:conf/mmm/ZhongFHWS19, zhong2019, rusu2019meta, pami20isif, ye2016person}, these approaches tend to break down in the low-data regimes due to the lack of sufficient labeled data for training. With the aim to learn new knowledge or concepts from a limited number of examples, few-shot learning approaches have been proposed to address the problem above~\cite{1597116, Sun2019Meta, Schonfeld_2019_CVPR, Kim_2019_CVPR, Chu_2019_CVPR, yao2019graph}. 

\begin{figure}[tp]
	\centerline{\includegraphics[scale = 0.47]{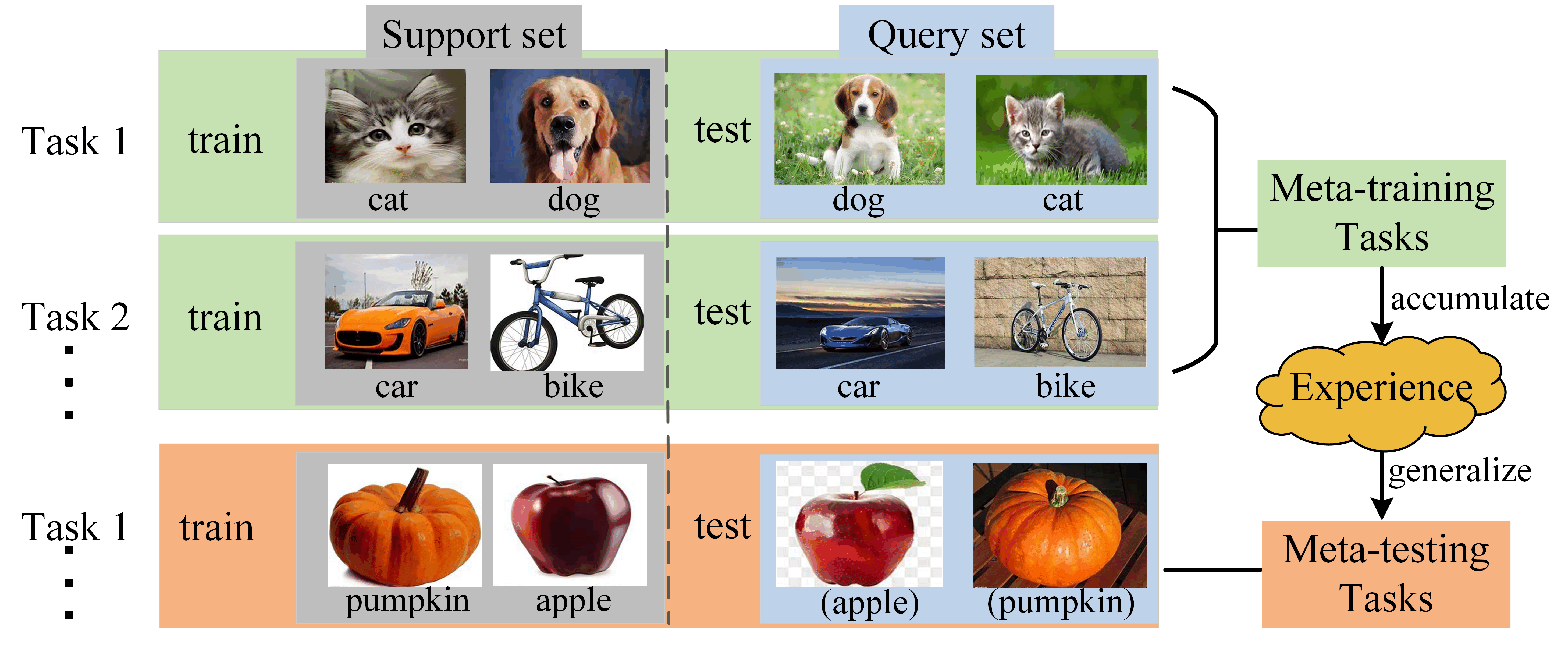}}
	\caption{An illustration of meta-learning for a few-shot learning task. The meta-learner is optimized with “experience” to adapt a base-learner to new tasks.}
	\label{problem_def}
\end{figure}

Meta-learning, as shown in Fig.~\ref{problem_def}, is one major kind of methods for few-shot learning recently. It contains a base-learner, and a meta-learner which adapts the base-learner to new tasks with few samples. The goal of meta-learning is accumulating ‘‘experience'' to learn a prior over tasks and generalize to various new tasks with very few training data. However, a key challenge for it is representation deficiency that using few samples to effectively calculate gradients in a high-dimensional parameter space is intractable in practice~\cite{rusu2019meta}. This results in the difficulty of representing common features and training the meta-learner well to generalize to new tasks.

\begin{figure*}[htbp]
	\centerline{\includegraphics[scale = 0.45]{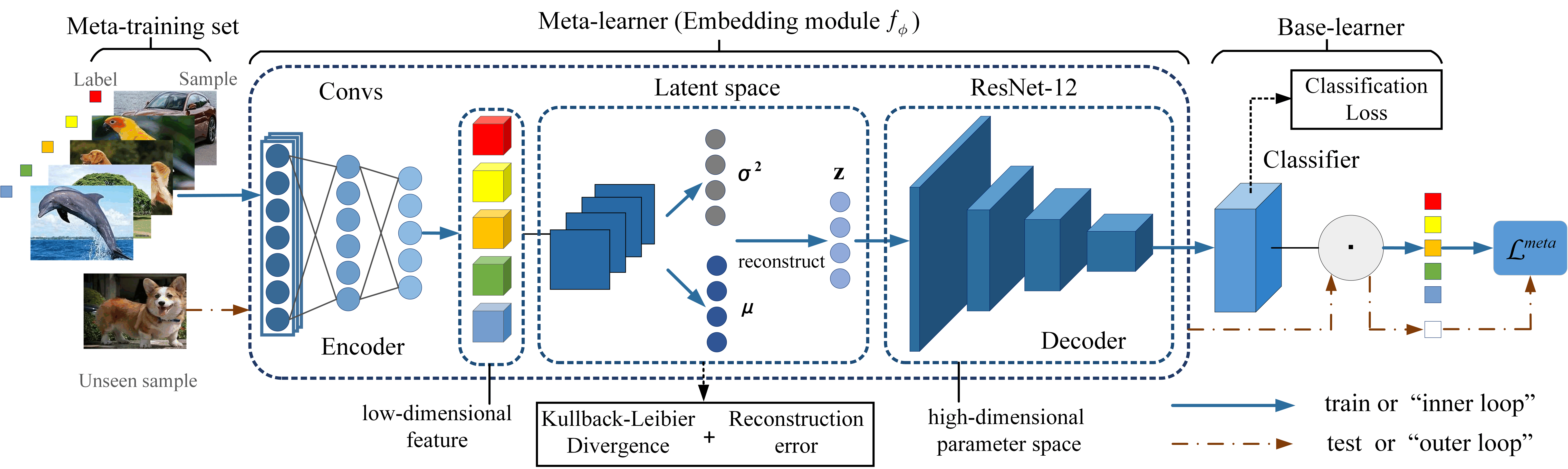}}
	\caption{Overview of the proposed method. The figure shows a typical 5-way 1-shot task based on our approach where the solid lines indicate the training process and the dashed lines indicate the test process during meta-training. Given an unseen sample from query set, the latent code \textbf{z} with additional information, reconstructed in latent space, contributes to better representations.}
	\label{overview}
\end{figure*}

Recently, a number of studies~\cite{finn2017model, vinyals2016matching, snell2017prototypical, PLATIPUS2018, bayesianMAML2018} in meta-learning seek to represent features directly with less expressive feature extractors. These methods generally alleviate the tendency of over-fitting in few-shot setting and the difficulty of representing features~\cite{Sun2019Meta, hui2019self}, caused by the representation deficiency, but at the expense of greater representation or expressiveness. Furthermore, the inherent task ambiguity~\cite{PLATIPUS2018} in few-shot learning also leads to difficulties when scaling to high-dimensional data. Therefore, it is desirable to develop an approach to achieve both better representation and ambiguity awareness.

In this paper, inspired by the out-performance of variational inference in generating extra information~\cite{kingma2014auto-encoding, DBLP:conf/cvpr/YeZYC19}, we propose a novel end-to-end meta-learning approach with complemented representations network (MCRNet) for few-shot image classification. Fig.~\ref{overview} overviews the proposed method. Specially, our approach embeds a latent space based on low-dimensional features learned from raw samples through encoding convolution layers. Then latent codes \textbf{z}, reconstructed in this space, are endued extra representation information to complement the representation deficiency and will be decoded in a high-dimensional parameter space for better representation. The gradient-based optimization is performed with the new loss in both latent space and high-dimensional parameter space. Besides, the latent space is established via variational inference with stochastic initialization, enabling the method to model the inherent task ambiguity in few-shot learning~\cite{PLATIPUS2018}. Our results demonstrate that MCRNet achieves state-of-the-art performance in classification tasks on three few-shot learning datasets including CIFAR-FS~\cite{bertinetto2019meta}, FC100~\cite{oreshkin2018tadam}, and miniImageNet~\cite{ravi2017optimization, vinyals2016matching}.

The main contributions of our work are threefold: 
\begin{itemize}
	\item We propose an end-to-end framework and interpolate a latent space to endue the reconstructed latent codes with more information, complementing the representation deficiency in a high-dimensional parameter space.
	\item The probabilistic latent space with stochastic initialization collaborates well with different base-learners and can be extended to other architecture with high-dimensional feature extractors in few-shot learning.
	\item We optimize the framework leveraging new loss function for the proposed latent space, which acquires better generalization across tasks and achieves the state-of-the-art performance in few-shot learning classification tasks.
\end{itemize}

\section{Related Work}
Adaptation to new information from limited knowledge is an important aspect of human intelligence, leading to the popularity of few-shot learning~\cite{finn2017model, vinyals2016matching, snell2017prototypical, sung2018learning, DBLP:journals/corr/abs-1904-06317}. Tackling the few-sample problems, one of the major few-shot learning methods is meta-learning, which aims to accumulate ‘‘experience'' and adapt the ‘‘experience'' to new tasks fast. Contemporary methods of meta-learning could be roughly classified into three categories. 1) Metric-based methods~\cite{snell2017prototypical, vinyals2016matching, sung2018learning}, which learn a similarity space through training similar metrics over the same class to enhance effectiveness. 2) Memory-based methods~\cite{oreshkin2018tadam, mishra2018a}, which use memory architecture to store key “experience” from seen samples and generalize to unseen tasks according to the stored knowledge. 3) Optimization-based methods~\cite{finn2017model, lee2018gradient, grant2018recasting}, which search for a suitable meta-learner that is conducive to fast gradient-based adaptation to new tasks. In this process, the meta-learner and base-leaner are continuously optimized in the “outer loop” and “inner loop”, and the optimization in our method is based on this concept.

MAML~\cite{finn2017model} is an important model in meta-learning and has been extended to many variants recently. Making use of learning Gaussian posteriors over parameters to model task ambiguity, \cite{PLATIPUS2018} and~\cite{bayesianMAML2018} propose probabilistic extensions to MAML. Unfortunately, these methods choose a less expressive architecture as a feature extractor, like 4CONV~\cite{finn2017model}, to avoid the representation deficiency in a high-dimensional parameter space. \cite{lee2019meta} tries to propose better base-learners and~\cite{hui2019self} interpolates self-attention relation network. These two extract features directly via more powerful architecture, ResNet-12~\cite{oreshkin2018tadam} or ResNet-101, but neither of them addresses the deficiency mentioned. LEO\cite{rusu2019meta}, using WRN-28-10~\cite{ZagoruykoK16}, takes the deficiency into account. However, the relation network incorporated in it may lead to data-dependency issue and the potential of latent space may be neglected, since it fails to perform optimization in high-dimensional parameter space.

In contrast, we learn a probabilistic latent space over model parameters to generate latent codes with extra representation information and perform adaptation with new loss in it and the high-dimensional parameter space. This enables our method to complement the representation deficiency in few-shot learning intuitively.

\section{METHODOLOGY}
\label{sectionMETHODOLOGY}
Our main goal is embedding a latent space, established via variational inference, to complement the representation deficiency so as to learn better representations for meta-learning.

\subsection{Primary problem definition}
We evaluate the proposed approach in the $\mathit{K}$-way, $\mathit{N}$-shot problem for few-shot image classification referring to the episodic formulation of~\cite{vinyals2016matching} and Fig.~\ref{problem_def}, where $\mathit{K}$ represents the number of classes for one task instance and $\mathit{N}$ represents the number of training samples per class. Typically, the value of $\mathit{N}$ is 1 or 5 and $\mathit{K}$ is 5.

The raw dataset is divided into 3 mutually disjoint meta-sets: meta-training set $\mathcal{S}^{train}$, meta-validation set $\mathcal{S}^{val}$, and meta-testing set $\mathcal{S}^{test}$, considering the model's generalizability assessment. $\mathcal{S}^{val}$ is used for model selection and $\mathcal{S}^{test}$ is used for the final evaluation. Each task/episode is constructed during meta-training. As shown in Fig.~\ref{problem_def}, data for one task $\mathcal T = (\mathcal{D}^{train}, \mathcal{D}^{test})$ are sampled from $\mathcal{S}^{train}$ as follows: the training/support set $\mathcal{D}^{train} = \{ (x_{n}^{k}, y_{n}^{k}) | k = 1, 2...\mathit{K}; n = 1, 2...\mathit{N}\}$ consists of $\mathit{K}$ classes selected from $S^{train}$ and $\mathit{N}$ samples per class. The test/query set $\mathcal{D}^{test}$ is composed of other different samples belonging to the same classes in $\mathcal{D}^{train}$. $\mathcal{D}^{test}$ plays the role of providing an estimate of generalization and further optimizing the meta-learning objective in each task. Notably, $\mathcal{S}^{test}$ should not be confused with $\mathcal{D}^{test}$.

\subsection{Variational inference for latent space construction}
Aiming to complement the representation deficiency, we apply a probabilistic scheme, based on~\cite{kingma2014auto-encoding}, to establish a latent space and reconstruct latent codes with additional information in it.

To implement the parametric generative model, we first use several convolution layers as the encoder. Assuming that we acquire raw input $x_{n}$, the latent code $z_{n}$ can be defined through a posterior distribution $p (z_{n}|x_{n})$ which is, however, usually hard to estimate. We therefore introduce $q_{\phi_{e}} (z_{n}|x_{n})$ with input data point $x_{n}$ and variational parameters $\phi_{e}$ to approximate this posterior distribution, as defined below: 
\begin{equation}
\begin{split}
q_{\phi_{e}} (z_{n}|x_{n}) = \mathcal N (z_{n}|\mu_{n}, \sigma_{n}^{2}I), 
\end{split}
\end{equation}
where the mean and standard deviation of the approximate posterior are denoted as $\mu_{n}$ and $\sigma_{n}$, which are obtained through an encoding neural network: 
\begin{equation}
\begin{cases}
\mu = W_{v}x + b_{v},\\
\log \sigma^{2} = W_{w}x + b_{w}, 
\end{cases}
\end{equation}
where the parameter group \{$W_{v}, W_{w}, b_{v}, b_{w}$\} contains the weights and biases of the neural network. We sample random variables from the posterior distribution $z^{l}_{n} \sim q_{\phi_{e}} (z_{n}|x_{n})$ to reconstruct the latent code: 
\begin{eqnarray}
z^{l}_{n}\! & = & \!f_{\phi_{e}} (x_{n}, \epsilon^{l}) \nonumber\\
& = & \!\mu_{n} + \sigma_{n}\odot\epsilon^{l}, 
\end{eqnarray}
subject to 
\[\epsilon^{l} \sim \mathcal N (0, 1), \]
where $f_{\phi_{e}}$ represents the production function of $z$ and $\odot$ denotes element-wise multiplication.

\begin{figure}[tbp]
	\centerline{\includegraphics[scale = 0.6]{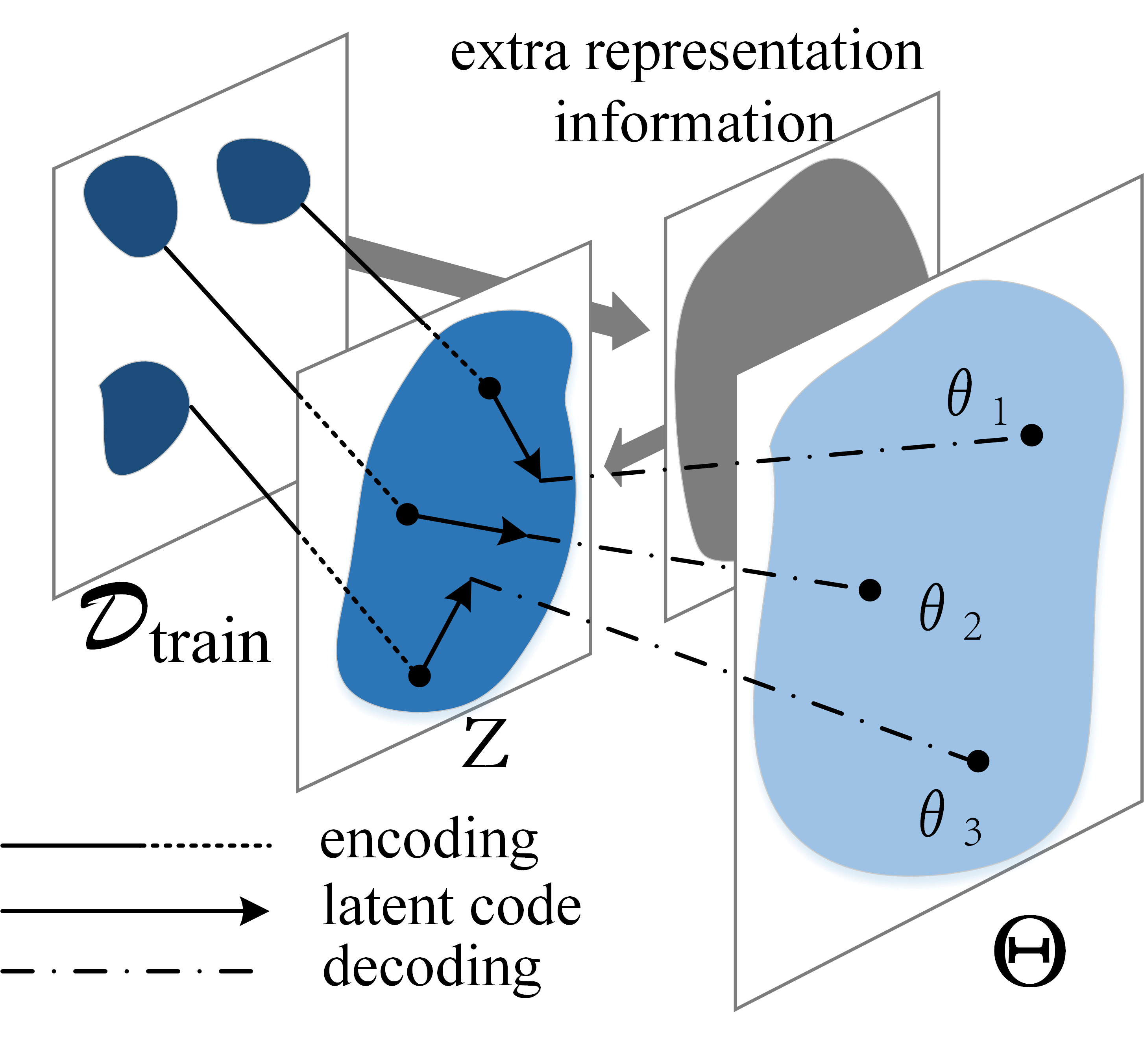}}
	\caption{Process of data transformation. The latent code is endued extra information, when reconstructed in latent space $\mathcal{Z}$, to complement the representation deficiency. It will be subsequently decoded in high-dimensional parameter space $\Theta$.}
	\label{latent}
\end{figure}

\begin{algorithm}
	\caption{MCRNet }
	\label{alg:A}
	$\textbf{Input}$: Meta-training set $S^{train}$; Encoding function $f_{\phi_{e}}$, and decoding function $f_{\phi_{d}}$; Learning rate $\lambda, \eta$ 
	\begin{algorithmic}[1]
		\STATE Initialize $\phi_{e}, \phi_{d}$ randomly
		\STATE Let $\phi = \{\phi_{e}, \phi_{d}, \lambda\}$
		\STATE \textbf{while} new batch \textbf{do}
		\STATE \hspace{2ex}\textbf{for} number of task in batch \textbf{do}
		\STATE \hspace{4ex}Sample task $\mathcal{T}_{i}$ = ($\mathcal{D}^{train}$, $\mathcal{D}^{test}$) from $S^{train}$
		\STATE \hspace{4ex}\textbf{for} number of sample in support set $\mathcal{D}^{train}$ \textbf{do}
		\STATE \hspace{6ex}Encode samples in $\mathcal{D}^{train}$ to $\mathbf{z}$ with complemented\\
		\hspace{6ex}representations using $f_{\phi_{e}}$
		\STATE \hspace{6ex}Decode $\mathbf{z}$ using $f_{\phi_{d}}$ 
		\STATE \hspace{6ex}Compute $\theta$ in base-learner using (6)
		\STATE \hspace{6ex}Compute training loss $\mathcal{L}_{\mathcal{T}_{i}}^{train}(\mathcal{D}^{train}; \phi, \theta, \varphi)$
		\STATE \hspace{6ex}Perform gradient optimization:\\
		\hspace{6ex}$\theta \gets \theta - \lambda\bigtriangledown_{\theta}\mathcal{L}_{\mathcal{T}_{i}}^{train}(\mathcal{D}^{train}; \phi, \theta, \varphi)$
		\STATE \hspace{4ex}\textbf{end for}
		\STATE \hspace{4ex}\textbf{for} number of sample in query set $\mathcal{D}^{test}$ \textbf{do}
		\STATE \hspace{6ex}Compute test loss $\mathcal{L}_{\mathcal{T}_{i}}^{test}(\mathcal{D}^{test}; \phi, \theta, \varphi)$
		\STATE \hspace{4ex}\textbf{end for}
		\STATE \hspace{4ex}Compute meta-training loss $\mathcal{L}^{meta}(\mathcal{S}^{train}; \theta, \phi)$
		\STATE \hspace{4ex}Perform gradient optimization:\\
		\hspace{4ex}$\phi \gets \phi - \eta\bigtriangledown_{\phi}\mathcal{L}^{meta}(\mathcal{S}^{train}; \theta, \phi)$
		\STATE \hspace{2ex}\textbf{end for}
		\STATE \textbf{end while}
	\end{algorithmic}
\end{algorithm}

Besides, in the method, we assume that both $p (z)$ and $q_{\phi_{e}} (z|x)$ follow a Multivariate Gaussian: 
\begin{equation}
p (\mathbf{z}) = \mathcal N (\mathbf{z}; 0, \mathbf{I}).
\end{equation}
To enforce that, we compute the variation loss as follows: 
\begin{eqnarray}
\label{eqn:variation-loss}
\begin{aligned}
\mathcal L_{var}\! = & \!-\!D_{KL}(q (z_{n}|x_{n})||p (z))\! + \! E_{q_{\phi_{e}}(x|z)}[\log p(x|z)]\\
\simeq & \frac{1}{2}\sum_{i = 1}^{I}(1 + \log ((\sigma^{(i)}_{n})^{2})-(\mu^{(i)}_{n})^{2} - (\sigma^{(i)}_{n})^{2})\\ 
& + \frac{1}{L}\sum_{l = 1}^{L}\log p(x_{n}|z_{n}^{l}), 
\end{aligned}
\end{eqnarray}
where the first term is Kullback-Liebler divergence, the second is reconstruction error and $\sigma^{(i)}_{n}$, $\mu^{(i)}_{n}$ respectively denote the $\mathit{i}$-th mean and standard deviation of feature samples.

In our method, as shown in Fig.~\ref{latent}, latent codes \textbf{z} are reconstructed through combining the original features $\mu$ and additional representation information $\sigma$ from the encoding neural network with variational inference. Meanwhile, extra information enables expressing the ambiguity in few-shot learning in a probabilistic manner~\cite{PLATIPUS2018}.

\subsection{Meta-learning}
For each task $\mathcal{T}_{i}$, the adaptation model produces a novel classifier $\mathcal{A}$, defined as the base-learner in meta-learning and optimized constantly. This process is performed on $\mathcal{D}^{train}$. The purpose of base-learner $\mathcal{A}$ is to estimate parameters $\theta$ of predictor for classification tasks. We obtain the parameter through minimizing the empirical loss: 
\begin{equation}
\label{eqn:empirical-loss}
\theta = \mathcal{A} (\mathcal{D}^{train}; \phi) = \mathop{\arg\min}_{\theta} \mathcal{L}^{base} (D^{train}; \theta, \phi) + \mathcal{C} (\theta), 
\end{equation}
where $\mathcal{L}^{base}$ is a loss function defined as negative log-likelihood of labels in our method and $\mathcal{C} (\theta)$ is designed as a regularization term. $\phi$ is parameter of embedding module $f_{\phi}$, as shown in Fig.~\ref{overview}.

As an integral part of meta-learning, the base-learner plays an important role and we choose the base-learners suggested in~\cite{lee2019meta}, considering that the objective of the base-learner in our method is also convex. There are two types of base-learners used in our method including Support Vector Machine (SVM) and Ridge Regression (RR), which are both based on multi-class linear classifiers and perform well as reported in~\cite{lee2019meta}. Specifically, a $\mathit{K}$-class linear base-learner could be expressed as $\theta$ = $\{w_{k}\}_{k = 1}^{K}$. As a result, (\ref{eqn:empirical-loss}) is transformed into the Crammer and Singer formulation~\cite{crammer2001algorithmic}: 
\begin{eqnarray}
\theta \! & = & \!\mathcal{A} (\mathcal{D}^{train}; \phi) \nonumber\\
\! & = & \!\mathop{\arg\min}_{\{w_{k}\}}\mathop{\min}_{\{\xi_{i}\}}\frac{1}{2} \sum_{k} \|w_{k}\|_{2}^{2} + C\sum_{n}\xi_{n}, 
\end{eqnarray}
subject to 
\[w_{y_{n}} \cdot f_{\phi} (x_{n}) - w_{k} \cdot f_{\phi} (x_{n}) \geq 1 - \delta_{y_{n}, k} - \xi_{n}, \forall n, k\]
where $\mathcal{D}^{train} = \{ (x_{n}, y_{n})\}$, $C$ is designed for regularization and $\delta_{y_{n}, k}$ is the Kronecker delta function.

Given the decoder parameters and base-learner, we then compute the ‘‘inner loop" loss in each episode as follows: 
\begin{equation}
\label{eqn:inner-loop-loss}
\begin{split}
& \mathcal{L}_{\mathcal{T}_{i}}^{train} (\mathcal{D}^{train}; \phi, \theta, \varphi) =\\& \sum_{ (x_{n}, y_{n})\in D^{train}_{i}}\![-\varphi w_{y_{n}}\!\cdot\!f_{\phi} (x_{n}) + \mathrm{\log}\!\sum_{k = 1}^{K}\mathrm{exp} (\varphi w_{k}\!\cdot\!f_{\phi} (x_{n}))], 
\end{split}
\end{equation}
where $\mathcal{T}_{i} = (\mathcal{D}^{train}_{i}, \mathcal{D}^{test}_{i})$ is the i-th sample from $\mathcal{S}^{train}$. $\varphi$ is a proportional optimizable parameter which has shown good performance in few-shot learning~\cite{bertinetto2019meta, oreshkin2018tadam} under conditions of SVM and RR as base-learners. Step 6 to 12 in Algorithm~\ref{alg:A} has shown the ‘‘inner loop".

After the ‘‘inner loop", we obtain a preliminary optimized base-learner and then conduct the second step ‘‘outer loop" on $\mathcal{D}^{test}$. In this part, encoding and decoding parameter $\phi_{e}$, $\phi_{d}$ will be updated to improve generalizability to unseen samples through gradient-based optimization. The ‘‘outer loop" is performed by minimizing the new loss $\mathcal{L}^{meta}$ consisting of two parts: 
\begin{equation}
\begin{split}
\mathcal{L}^{meta} (\mathcal{S}^{train}; \theta, \phi)\! = \!\!\sum_{\mathcal{T}_{i}\sim S^{train}}\![\mathcal{L}_{\mathcal{T}_{i}}^{test} (\mathcal{D}^{test}; \phi, \theta, \varphi)\! + \!\beta \mathcal {L}_{var}]
\end{split}
\end{equation}
where the first term is the deformation of (\ref{eqn:inner-loop-loss}) on $\mathcal{D}^{test}$ and the second term uses a variation loss with optimizable weight $\beta$ to regularize the latent space which is defined in (\ref{eqn:variation-loss}). Step 4 to 18 in Algorithm~\ref{alg:A} has shown the ‘‘outer loop".

Once finishing the meta-training, we evaluate the generalizability of model to unseen data tuple $\mathcal{S}_{j} = (\mathcal{D}^{train}_{j}, \mathcal{D}^{test}_{j})$, which is sampled from $S^{test}$. Hence, we compute the loss for evaluation during meta-testing as follows: 
\begin{equation}
\mathcal{L}^{meta} (\mathcal{S}^{test}; \theta, \phi).
\end{equation}
We use the meta-testing set $\mathcal{S}^{test}$ for evaluation, and the parameters of the model will not be updated in each episode $\mathcal{S}_{j}$.
\begin{table*}[tbp]
	\centering
	\setlength{\tabcolsep}{2mm}
	\fontsize{10.0}{9.0}\selectfont
	\begin{threeparttable}
		\caption{Comparisons of average classification accuracy (\%) with 95\% confidence intervals on the CIFAR-FS and FC100. ‘‘SVM'' or ‘‘RR'' means using SVM or Ridge Regression as base-learner.}
		\label{table1}
		\begin{tabular}{lccccc}
			\toprule
			\multirow{2}{*}{method} & \multirow{2}{*}{backbone} & \multicolumn{2}{c}{CIFAR-FS} & \multicolumn{2}{c}{FC100}\cr
			\cmidrule (lr){3-4} \cmidrule (lr){5-6}
			& & 1-shot & 5-shot & 1-shot & 5-shot\cr
			\midrule
			Relation Networks~\cite{sung2018learning} & 4CONV & 55.0 $\pm$ 1.0 & 69.3 $\pm$ 0.8 & - & -\cr
			Prototypical Networks~\cite{snell2017prototypical} & 4CONV & 55.5 $\pm$ 0.7 & 72.0 $\pm$ 0.6 & 35.3 $\pm$ 0.6 & 48.6 $\pm$ 0.6\cr
			MAML~\cite{finn2017model} & 4CONV & 58.9 $\pm$ 1.9 & 71.5 $\pm$ 1.0 & - & -\cr
			R2D2~\cite{bertinetto2019meta} & 4CONV & 65.3 $\pm$ 0.2 & 79.4 $\pm$ 0.1 & - & -\cr
			Fine-tuning~\cite{DBLP:journals/corr/abs-1909-02729} & ResNet-12 & 64.66 $\pm$ 0.73 & 82.13 $\pm$ 0.50 & 37.52 $\pm$ 0.53 & 55.39 $\pm$ 0.57\cr	
			TADAM~\cite{oreshkin2018tadam} & ResNet-12 & - & - & 40.1 $\pm$ 0.4 & 56.1 $\pm$ 0.4\cr
			MTL~\cite{Sun2019Meta} & ResNet-12$^{\P}$ & - & - & {\bf43.6 $\pm$ 1.8} & 55.4 $\pm$ 0.9\cr
			\midrule
			Baseline-RR~\cite{lee2019meta} & ResNet-12 & 72.6 $\pm$ 0.7 & 84.3 $\pm$ 0.5 & 40.5 $\pm$ 0.6 & 55.3 $\pm$ 0.6\cr
			Baseline-SVM~\cite{lee2019meta} & ResNet-12 & 72.0 $\pm$ 0.7 & 84.2 $\pm$ 0.5 & 41.1 $\pm$ 0.6 & 55.5 $\pm$ 0.6\cr
			\midrule
			MCRNet-RR (ours) & ResNet-12 & {\bf73.8 $\pm$ 0.7} & {\bf85.2 $\pm$ 0.5} & 40.7 $\pm$ 0.6 & {\bf56.6 $\pm$ 0.6}\cr
			MCRNet-SVM (ours) & ResNet-12 & {\bf74.7 $\pm$ 0.7} & {\bf86.8 $\pm$ 0.5} & 41.0 $\pm$ 0.6 & {\bf57.8 $\pm$ 0.6}\cr
			\bottomrule
		\end{tabular}
		{$^{\P}$ indicates that the method is not end-to-end.}
	\end{threeparttable}
\end{table*}

\begin{table*}[htbp]
	\setlength{\tabcolsep}{5.0mm}
	\centering
	\fontsize{10.0}{9.0}\selectfont
	\begin{threeparttable}
		\caption{Comparisons of average classification accuracy (\%) with 95\% confidence intervals on the miniImageNet.}
		\label{table2}
		\begin{tabular}{lccc}
			\toprule
			method & backbone & 1-shot & 5-shot\cr
			\midrule
			Meta-Learning LSTM~\cite{ravi2017optimization} & 4CONV & 43.44 $\pm$ 0.77 & 60.60 $\pm$ 0.71\cr
			Matching networks~\cite{vinyals2016matching} & 4CONV & 43.56 $\pm$ 0.84 & 55.31 $\pm$ 0.73\cr
			MAML~\cite{finn2017model} & 4CONV & 48.70 $\pm$ 1.84 & 63.11 $\pm$ 0.92\cr
			Prototypical Networks~\cite{snell2017prototypical} & 4CONV & 49.42 $\pm$ 0.78 & 68.20 $\pm$ 0.66\cr
			Relation Networks~\cite{sung2018learning} & 4CONV & 50.44 $\pm$ 0.82 & 65.32 $\pm$ 0.70 \cr
			R2D2~\cite{bertinetto2019meta} & 4CONV & 51.2 $\pm$ 0.6 & 68.8 $\pm$ 0.1\cr
			SRAN~\cite{hui2019self} & ResNet-101$^{\P}$ & 51.62 $\pm$ 0.31 & 66.16 $\pm$ 0.51\cr
			DN4~\cite{DBLP:journals/corr/abs-1903-12290} & ResNet-12 & 54.37 $\pm$ 0.36 & 74.44 $\pm$ 0.29\cr
			SNAIL~\cite{mishra2018a} & ResNet-12 & 55.71 $\pm$ 0.99 & 68.88 $\pm$ 0.92\cr
			Fine-tuning~\cite{DBLP:journals/corr/abs-1909-02729} & ResNet-12 & 56.67 $\pm$ 0.62 & 74.80 $\pm$ 0.51\cr
			TADAM~\cite{oreshkin2018tadam} & ResNet-12$^{\P}$ & 58.50 $\pm$ 0.30 & 76.70 $\pm$ 0.30\cr
			CAML~\cite{DBLP:conf/iclr/JiangHVCCM19} & ResNet-12 & 59.23 $\pm$ 0.99 & 72.35 $\pm$ 0.71\cr
			TPN~\cite{DBLP:journals/corr/abs-1805-10002} & ResNet-12 & 59.46 & 75.65\cr
			wDAE-GNN~\cite{DBLP:conf/cvpr/GidarisK19} & WRN-28-10 & 61.07 $\pm$ 0.15 & 76.75 $\pm$ 0.11\cr
			MTL~\cite{Sun2019Meta} & ResNet-12$^{\P}$ & 61.2 $\pm$ 1.8 & 75.5 $\pm$ 0.8\cr
			LEO~\cite{rusu2019meta} & WRN-28-10$^{\P}$ & 61.76 $\pm$ 0.08 & 77.59 $\pm$ 0.12\cr
			\midrule
			LEO$^{\ast}$~\cite{rusu2019meta} & ResNet-12$^{\P}$ & 58.67 $\pm$ 0.07$^{\ast}$ & 73.45 $\pm$ 0.12$^{\ast}$\cr
			Baseline-RR~\cite{lee2019meta} & ResNet-12 & 60.02 $\pm$ 0.64$^{\ast}$ & 76.51 $\pm$ 0.49$^{\ast}$\cr
			Baseline-SVM~\cite{lee2019meta} & ResNet-12 & 60.73 $\pm$ 0.65$^{\ast}$ & 76.16 $\pm$ 0.49$^{\ast}$\cr
			\midrule
			MCRNet-RR (ours) & ResNet-12 & 61.32 $\pm$ 0.64 & {\bf78.16 $\pm$ 0.49}\cr
			MCRNet-SVM (ours) & ResNet-12 & {\bf62.53 $\pm$ 0.64} & {\bf80.34 $\pm$ 0.47}\cr
			\bottomrule
		\end{tabular}
		{$^{\P}$ indicates those methods are not end-to-end. $^{\ast}$ indicates those methods that are reproduced by ourselves for comparison of convergence.}
	\end{threeparttable}
\end{table*}

\section{Experiments}
We evaluate the proposed MCRNet for few-shot classification tasks on the unseen meta-testing set and compare it with the state-of-the-art methods.

\subsection{Datasets}
\textbf{CIFAR-FS} is a new standard benchmark for few-shot learning tasks, consisting of 100 classes from CIFAR-100~\cite{krizhevsky2010cifar}. These classes are randomly divided into 64, 16, and 20 classes for meta-training, meta-validation and meta-testing with 600 images of size 32 $\times$ 32 in each class.

\textbf{FC100} is another dataset derived from CIFAR-100 and similar to CIFAR-FS. Differently, its 100 classes are grouped into 20 advanced classes and divided into 60 classes from 12 advanced classes for meta-training, 20 classes from 4 advanced classes for meta-validation, and 20 classes from 4 advanced classes for meta-testing. This dataset is intended to reduce the semantic similarity between classes.

\textbf{MiniImageNet} is a common dataset used for few-shot learning tasks with 100 classes randomly sampled from ILSVRC-2012~\cite{russakovsky2015imagenet} and divided into meta-training, meta-validation, and meta-testing sets with 64, 16, and 20 classes respectively. Each class contains 600 images of size 84 $\times$ 84, and we choose commonly-used class split in~\cite{ravi2017optimization}.

\begin{figure*}[htbp]
	\centering
	\subfigure[]{
		\begin{minipage}[t]{0.50\linewidth}
			\centering
			\includegraphics[width = 2.5in, height = 1.9in]{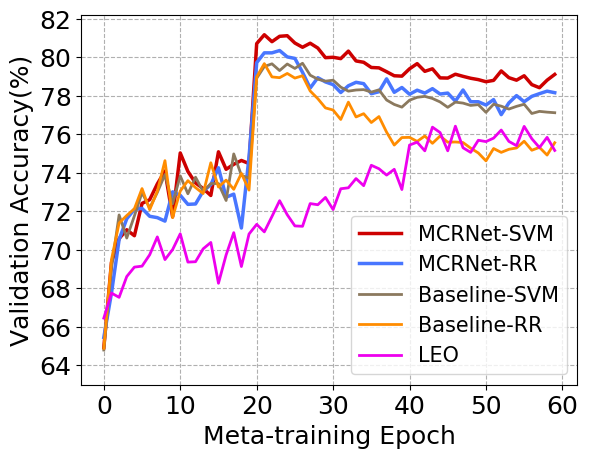}
		\end{minipage}%
	}%
	\subfigure[]{
		\begin{minipage}[t]{0.50\linewidth}
			\centering
			\includegraphics[width = 2.5in, height = 1.9in]{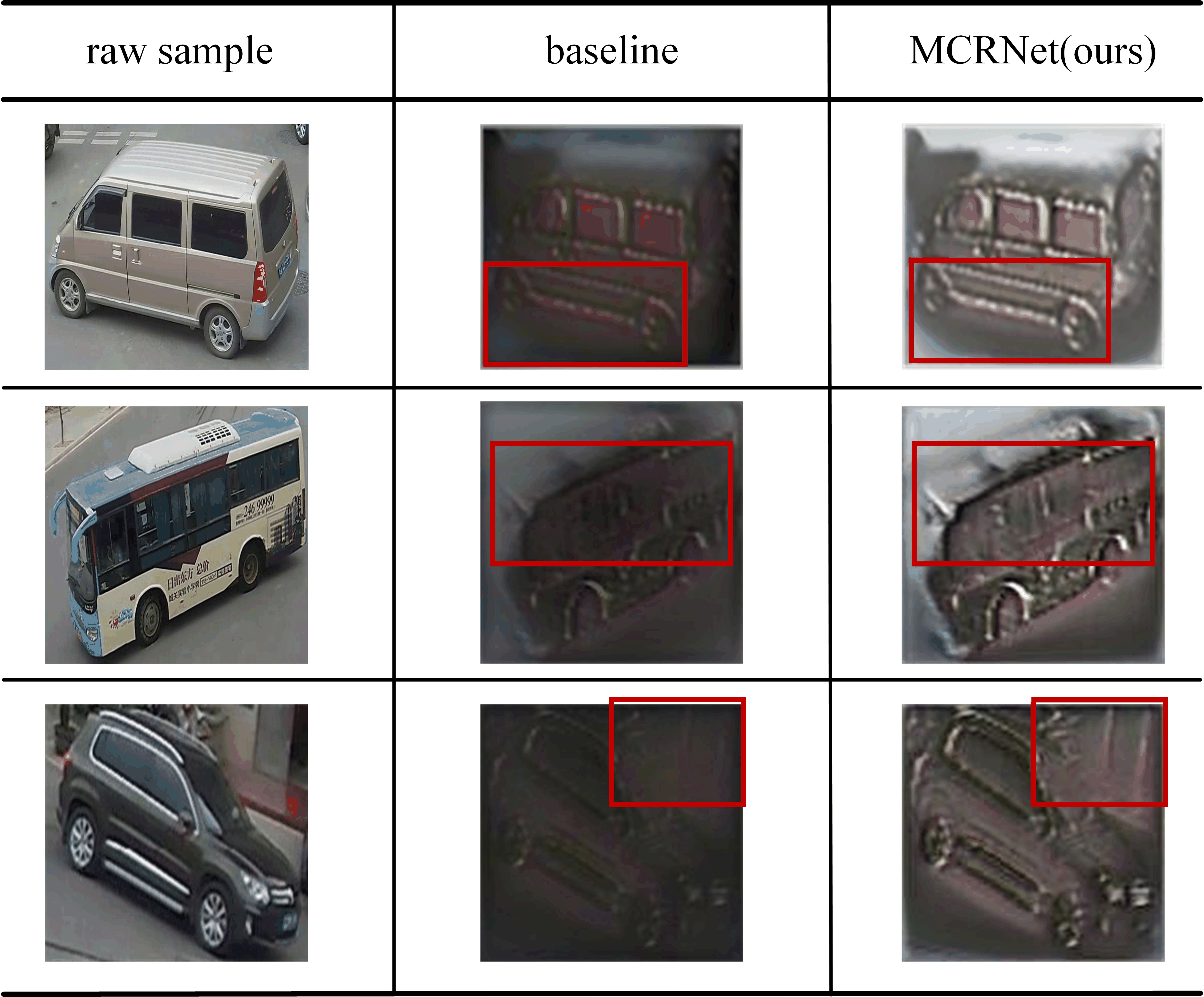}
		\end{minipage}%
	}%
	\caption{(a) Comparison of convergence on miniImageNet. Accuracies are obtained on meta-validation set after each epoch. (b) Some examples of the representation output from baseline and MCRNet.}
	\label{convergence}
\end{figure*}

\subsection{Network architecture}
In a nutshell, our method consists of meta-learner (including an encoder and a decoder), and base-learner. The encoder is composed of several convolution layers and sampling architecture, which is utilized to produce latent codes. The dimension of the latent space is determined by the encoder. Currently, there are two popular network models for the decoder to learn high-dimensional features in meta-learning, 4CONV and ResNet-12. Both of them consist of 4 blocks with several convolutions, Batch Normalization and pooling layers, but differ in the number of filters in each block. We choose ResNet-12 as the decoder in our method, considering the poor performance of 4CONV for few-shot learning~\cite{Sun2019Meta}. For the base-learner, we utilize two classic linear methods, multi-class SVM and RR.

In our experiment, the hyperparameters generally remain consistent with the baseline~\cite{lee2019meta} for fair comparisons. We apply stochastic gradient descent-based optimization and set the Nesterov momentum and weight decay to 0.9 and 0.0005, respectively. Totally, we meta-train the model for 60 epochs with 1000 episodes in each epoch. Besides, we set the learning rate to 0.1, 0.006, 0.0012, and 0.00024 at epochs 1, 20, 40, and 50 respectively.

\subsection{Results and analysis}
\subsubsection{\textbf{Few-shot image classification}}
Our experimental results on the datasets are averaged over 1000 test episodes and summarized in TABLE~\ref{table1} and TABLE~\ref{table2}. In general, compared to the baseline and current state-of-the-art methods, the accuracies of our model in 5-way 1- and 5-shot classification tasks both get improved. 

As detailed in TABLE~\ref{table1}, in 1-shot and 5-shot test, our method achieves 10.04\% and 4.67\% improvement over Fine-tuning~\cite{DBLP:journals/corr/abs-1909-02729} on CIFAR-FS dataset. On FC100, compared to the baseline, our method gets improved by 2.3\% in 5-shot classification tasks. TABLE~\ref{table2} illustrates our experiment on miniImageNet dataset. Our method beats the baseline accuracy by 1.80\% and 4.18\% in 1-shot and 5-shot test respectively. On miniImageNet, the proposed MCRNet still achieves 4.84\% and 2.75\% improvement over the competitive method MTL~\cite{Sun2019Meta} and LEO~\cite{rusu2019meta} in 5-shot classification respectively.



Our method provides a solution to break the representation limit in a high-dimensional parameter space through complementing the representation deficiency, and further fulfills the potential of ResNet-12 in representation learning. Intuitively, it outperforms those with ResNet-12, especially on CIFAR-FS and miniImageNet. Besides, collaborating well with different base-learners, MCRNet outperforms existing models on different datasets, revealing the good applicability of our method.

Although the methods in~\cite{hui2019self, rusu2019meta, DBLP:conf/cvpr/GidarisK19, Sun2019Meta} use pre-trained or deeper feature extractors and TADAM~\cite{oreshkin2018tadam} co-trains the feature extractor on 5- and 64-way classification tasks to tackle the representation limit in few-shot learning, our method still achieves improvement generally. Besides, our model is meta-trained end-to-end, allowing us to clearly observe the impact of latent space for meta-learning. Obviously, it is essential to complement the representation deficiency when using an expressive architecture in few-shot learning.

\subsubsection{\textbf{Ablation study}}
We reproduce the baseline~\cite{lee2019meta} and the state-of-the-art LEO~\cite{rusu2019meta} with the provided code on miniImageNet using ResNet-12 and compare with them in terms of convergence rate. As shown in Fig.~\ref{convergence}(a), the number of tasks required to reach convergence is almost the same as the baseline, but our method achieves higher accuracy. LEO needs a pre-trained raw embedding network. It adopts MAML as its training methods, and the relation network is embedded in it. Therefore, it is hard for LEO to train the whole model well and more iterations are needed to reach convergence. As shown in Table~\ref{table2}, compared to the LEO with ResNet-12, the accuracies of our method get improved by 3.86\% and 6.69\% in 5 way 1- and 5-shot classification on miniImageNet respectively. 

Fig.~\ref{convergence}(b) shows some examples of representation output from baseline and our model. As we can see from the red box, our model is able to learn more details, thanks to the complemented representations. Compared to the baseline, the representation outputs generated from our model tend to perform better, such as higher brightness and sharper texture, which is beneficial to better generalization.

In regard to the influence of the latent space dimension on the model's performance, as shown in TABLE~\ref{table3}, we found that for a particular dataset and base-learner, there is a suitable latent space dimension that contributes to the improvement of generalizability. In our experiments, this value is mostly 64, which means the huge potential of our method in the high latent dimension. Remarkably, with dimension 64, our method achieves about 4\% improvement in 5-way classification accuracy on miniImageNet. However, considering the interface with ResNet-12 and fair comparison, the dimension setting in our experiment is up to 64.

\begin{table}[tbp]
	\centering
	\setlength{\tabcolsep}{1.0mm}
	\fontsize{9.0}{8.5}\selectfont
	\begin{threeparttable}
		\caption{Comparisons of average classification accuracy (\%) in different dimensions on the CIFAR-FS, FC100, and miniImageNet.}
		\label{table3}
		\begin{tabular}{cccccccc}
			\toprule
			\multirow{2}{*}{Latent Dimension} & \multirow{2}{*}{shot} & \multicolumn{2}{c}{CIFAR-FS} & \multicolumn{2}{c}{FC100} & \multicolumn{2}{c}{miniImageNet}\cr
			\cmidrule (lr){3-4} \cmidrule (lr){5-6} \cmidrule (lr){7-8}
			& & RR & SVM & RR & SVM & RR & SVM\cr
			\midrule
			without MCRNet&1&72.6&72.0&40.5&{\bf41.1}&60.0&60.7\cr
			8&1&69.4&70.1&39.0&39.8&57.2&58.7\cr
			16&1&70.3&71.2&39.2&40.4&58.4&60.1\cr
			32&1&72.9&72.5&{\bf40.7}&40.7&60.3&61.5\cr
			64&1&{\bf73.8}&{\bf74.7}&40.4&41.0&{\bf61.3}&{\bf62.5}\cr
			\midrule
			without MCRNet&1&84.3&84.2&55.3&55.5&76.5&76.2\cr
			8&5&82.6&83.8&53.5&54.0&75.8&75.4\cr
			16&5&83.8&84.3&54.0&55.4&76.4&77.5\cr
			32&5&84.5&85.5&{\bf56.6}&56.6&76.9&79.0\cr
			64&5&{\bf85.2}&{\bf86.8}&55.4&{\bf57.8}&{\bf78.2}&{\bf80.3}\cr
			\bottomrule
		\end{tabular}
	\end{threeparttable}
\end{table}

\section{Conclusion}
In this paper, we proposed an MCRNet for few-shot learning, which achieved state-of-the-art performance on the challenging 5-way 1- and 5-shot CIFAR and ImageNet classification problems. The MCRNet made use of a probabilistic framework to learn a latent space to complement the representation deficiency with extra representation information and broke the representation limit in a high-dimensional parameter space, resulting in better generalization across tasks. The experimental results also demonstrated the good performance of our method when applying to different base-learners.

\section*{Acknowledgment}
This work was supported in part by Fundamental Research Funds for the Central Universities under Grant 191010001, and Hubei Key Laboratory of Transportation Internet of Things under Grant 2018IOT003, Department of Science and Technology of Hubei Provincial People's Government under Grant 2017CFA012, and Ministry of Science and Technology of Taiwan under Grants MOST 108-2634-F-007-009.




\bibliographystyle{IEEEtran}
\bibliography{strings, refs}
%

\end{document}